\documentclass[11pt,a4paper]{article}
\usepackage[hyperref]{emnlp-ijcnlp-2019}
\usepackage[T1]{fontenc}
\usepackage{times}
\usepackage{latexsym}
\usepackage{adjustbox}
\usepackage{amsmath}
\usepackage{amssymb}
\usepackage{booktabs}
\usepackage{enumitem}
\usepackage{mathtools}
\usepackage{tikz}
\usepackage{pgfplots}
\usepackage{url}
\usepackage{siunitx}

\usetikzlibrary{fit}

\setlist[description]{%
  labelindent=0em,
  leftmargin=1.5em,
  style=nextline
}

\aclfinalcopy

\DeclarePairedDelimiter\parens{\lparen}{\rparen}
\newcommand\softmax[1]{\operatorname{softmax}\parens*{#1}}
\newcommand\LSTM[1]{\operatorname{LSTM}\parens*{#1}}
\newcommand\avg[1]{\operatorname{avg}\parens*{#1}}
\DeclareMathOperator*{\argmax}{arg\,max}

\tikzset{
  amr/.style={font=\small\ttfamily},
  main/.style={draw,inner sep=2.5pt,minimum size=10pt}
}
\tikzset{>=latex}
\pgfplotsset{compat=1.14}

\title{AMR-to-Text Generation with Cache Transition Systems}

\author{Lisa Jin \and Daniel Gildea \\
  Department of Computer Science \\
  University of Rochester \\
  {\tt \{lisajin,gildea\}@cs.rochester.edu } \\}

\date{}

\begin{document}
\maketitle
\begin{abstract}
  Text generation from AMR involves emitting sentences that reflect the meaning of their AMR annotations. Neural sequence-to-sequence models have successfully been used to decode strings from flattened graphs (e.g., using depth-first or random traversal). Such models often rely on attention-based decoders to map AMR node to English token sequences. Instead of linearizing AMR, we directly encode its graph structure and delegate traversal to the decoder. To enforce a sentence-aligned graph traversal and provide local graph context, we predict transition-based parser actions in addition to English words. We present two model variants: one generates parser actions prior to words, while the other interleaves actions with words.
\end{abstract}

\section{Introduction}
Abstract Meaning Representation or AMR \cite{banarescu2013abstract} is a directed graph of labeled concepts and relations that captures sentence semantics. The propositional meaning behind its concepts abstracts away lexical properties. AMR is tree-like in structure as it has a single root node and few reentrancies, or children with multiple parents. The goal of AMR-to-text generation is to recover the original sentence realization given an AMR. This task can be seen as the reverse of the structured prediction found in AMR parsing.

In contrast to parsing, the NLG problem involves mapping an oftentimes sparse AMR graph onto an English sentence---a source of ambiguity. The lack of gold AMR-sentence pairs further increases a model's risk of overfitting.

\begin{figure}
  \centering
  \begin{tikzpicture}
    \draw[amr] (3,3.5) node{open-01};
    \draw[amr] (1,2) node{center};
    \draw[amr] (3,2) node{date-entity};
    \draw[amr] (5,2) node{formal};
    \draw[amr] (3,0.5) node{2009};
    \draw [->] (2.75,3.25) -- (1,2.25);
    \draw [->] (3,3.25) -- (3,2.25);
    \draw [->] (3.25,3.25) -- (5,2.25);
    \draw [->] (3,1.75) -- (3,0.75);
    \draw[amr] (1.5,3) node{ARG1};
    \draw[amr] (3.5,2.6) node{time};
    \draw[amr] (4.5,3) node{manner};
    \draw[amr] (3.5,1.25) node{year};
  \end{tikzpicture}
\caption{An example AMR annotating the sentence, ``The center will formally open in 2009''.}
\label{fig:sample-amr}
\vspace{-1em}
\end{figure}
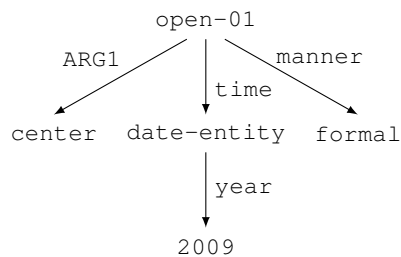

Previous work with encoder-decoder models used a soft attention mechanism to learn weights over input encodings per decoding step \cite{bahdanau2015neural}. This general architecture has been extended to the AMR-to-text task with various modifications to the encoder. For example, \citet{konstas2017neural} apply the sequential encoder on a depth-first linearized AMR after simplifying the graph structure through anonymization. \citet{song2018graph} instead use a recurrent graph encoder to directly model the graph structure.

Though soft attention allows the decoder full access to the input graph over time, NLG models may benefit from a more sequential form of attention. Since parsing graphs from sentences relies on input word order to construct AMR nodes, a similar dependency may exist in the text generation direction. One method that could formalize the relationship between graph structure and word order is transition-based parsing. Given a sentence-aligned vertex sequence, a parser builds the output AMR one vertex at a time. Thus, generation can be strictly defined as reverse parsing: processing the AMR in word order and emitting each concept-aligned English span.

Models predicting parser actions already show promising performance in parsing. To map sentences to parser actions, \citet{dyer2015transition} maintain continuous representations of transition system state over parsing steps, while \citet{buys2017robust} use hard attention on sentence token encodings. Conceivably, parser configuration can bolster prediction of word-order AMR traversal in NLG, just as in parsing \cite{peng2018sequence}.

We use the cache transition system \cite{gildea2018cache} that extends stack-based methods to semantic graph outputs. To the typical buffer and stack, this system adds a fixed size cache of vertices between which edges may be built. The cache's size is an upper-bound on the treewidth, or maximum size subgraph in an optimal tree decomposition, of graphs it can parse.

Two NLG decoders are presented, differing in their use of cache transition parsing actions. The \textit{action-conditioned} decoder (Section \ref{sec:ac}) predicts parser actions to process AMR in word order, then concept-wise English spans. In contrast, the \textit{joint action-word} decoder (Section \ref{sec:jaw}) alternates between action and span prediction, merging the two sequences to provide a shared history. It allows parser actions access to English spans aligned to AMR concepts from earlier parsing steps.

\section{Generation in Terms of Parsing}
Here we describe how parsing actions can apply to both $\mathbf{w} \mapsto G$ in parsing and the reverse $G \mapsto \mathbf{w}$ in NLG for string $\mathbf{w}$ and graph $G = (V, E)$. Parsing can be formulated as $G = f(g(\mathbf{w}))$ and generation as $\mathbf{w} = f^\prime(g^\prime(G))$, where
\begin{equation*}
  \begin{aligned}[c]
  g&\colon \mathbf{w} \mapsto \pi_\beta,\\
  f&\colon \pi_\beta \mapsto G,
\end{aligned}
\quad\quad
\begin{aligned}[c]
  g^\prime&\colon G \mapsto (\pi_\beta, E),\\
  f^\prime&\colon (\pi_\beta, E) \mapsto \mathbf{w}.
\end{aligned}
\end{equation*}

During parsing, the \textit{concept identification} stage $g(\cdot)$ maps $\mathbf{w}$ to a sequence of vertices $\pi_\beta$. Using JAMR \cite{flanigan2014discriminative} alignments, we assume a many-to-one mapping from spans of $\mathbf{w}$ to concepts of $G$. The \textit{edge identification} stage $f(\cdot)$ parses vertices $\pi_\beta$ to form connected graph $G$.

\begin{table*}
  \small
  \centering
  \begin{tabular*}{0.85\textwidth}{rccclc}
    stack & cache & buffer & edges & word span & preceding action\\
    \midrule
    {\ }[] & [\$, \$, \$] & \{o, c, d, f, 2\} & \{A, t, m, y\} & --- & --- \\{}
    [1, \$] & [\$, \$, c] & \{o, d, f, 2\} & \{A, t, m, y\} & the center will & $\textit{Push}(\text{c}, 1)$ \\{}
    [1, \$, 1, \$] & [\$, c, f] & \{o, d, 2\} & \{A, t, m, y\} & formally & $\textit{Push}(\text{f}, 1)$ \\{}
    [1, \$, 1, \$, 1, \$] & [c, f, o] & \{d, 2\} & \{t, y\} & open in & $\textit{Push}(\text{o}, 1)$ \\{}
    [1, \$, 1, \$, 1, \$, 1, c] & [f, o, d] & \{2\} & \{y\} & --- & $\textit{Push}(\text{d}, 1)$ \\{}
    [1, \$, 1, \$, 1, \$, 1, c, 1, f] & [o, d, 2] & \{\} & \{\} & 2009 . & $\textit{Push}(\text{2}, 1)$ \\{}
    [1, \$, 1, \$, 1, \$, 1, c] & [f, o, d] & \{\} & $\{\}$ & --- & \textit{Pop} \\{}
    [1, \$, 1, \$, 1, \$] & [c, f, o] & \{\} & \{\} & --- & \textit{Pop} \\{}
    [1, \$, 1, \$] & [\$, c, f] & \{\} & \{\} & --- & \textit{Pop} \\{}
    [1, \$] & [\$, \$, c] & \{\} & \{\} & --- & \textit{Pop} \\{}
    [] & [\$, \$, \$] & \{\} & \{\} & --- & \textit{Pop} \\{}
  \end{tabular*}
  \caption{Example run of cache transition system on Figure \ref{fig:sample-amr} AMR with cache size $k = 3$. Vertices and edges are denoted by the first letters of their labels (e.g., o $\gets$ \texttt{open-01}, A $\gets$ \texttt{ARG0}).}
  \label{tab:sample-run}
\end{table*}

To recover surface form $\mathbf{w}$ from $G$ for generation, the above process can be applied analogously. The buffer is now initialized with unordered vertices $V$. Since $G$ is given, we can shift vertices $V$ according to order $\pi_\beta$ using edges $E$ as guidance in $g^\prime(\cdot)$. The function $f^\prime(\cdot)$ expands each concept $v \in \pi_\beta$ into English spans concatenated to form $\mathbf{w}$. Our two decoder variants capture the effects of applying $g^\prime(\cdot)$ followed by $f^\prime(\cdot)$ or interleaving them.

\section{Cache Transition Parser}
We first introduce the cache transition system by \citet{gildea2018cache} in terms of parsing, then discuss how we simplify it to suit the NLG task.

\subsection{Background}
A cache transition parser is composed of three data structures: buffer, cache, and stack. The buffer is initialized with $n$ ordered vertices, and the cache is fixed to a predefined size \textit{k}. A configuration of the parser is denoted $$C = (\beta, \eta, \sigma, G_p),$$ where the first three elements represent the aforementioned data structures and $G_p$ is the partial graph at a point in the parsing process.

The parser operates by iteratively moving vertices from the buffer to the cache, where edges between cache elements are formed until the buffer is empty and target graph $G$ is parsed. Cache vertices can also be pushed to or popped from the stack. Specifically, cache vertex $\eta[i] = u$ can be displaced to the top of the stack by incoming buffer vertex $\beta[0] = v$. After this action, $\beta[j] \coloneqq \beta[j - 1], \forall j < |\beta|$, $\eta[k] \coloneqq v$, and $\sigma[-1] \coloneqq (i, u)$ with $\sigma[-k - 1] \coloneqq \sigma[-k], \forall k \in [1, |\sigma|]$. After all edges of $v$ have been formed, $u$ can be popped from the stack and assume its original position $\eta[i]$ with $v$ evicted from the cache at $\eta[k]$.

The parser is governed by the following actions.
\begin{enumerate}
  \item \textit{Shift} signals that the parser is done building edges within the current cache and can focus on the next buffer element $\beta[0]$.
  \item \textit{PushIndex(i)} moves the front buffer element to the cache at $\eta[k]$ and pushes the previous $\eta[i]$ to the stack using tuple $(i, \eta[i])$.
  \item \textit{Arc(i, d, l)} augments partial graph $G_p$ by connecting the rightmost cache element $\eta[k]$ to $\eta[i]$ with a directed, labeled edge $(d, l)$.
  \item \textit{Pop} restores the cache to its state prior to the \textit{PushIndex(i)} acting on the topmost stack element $\sigma[-1]$.
\end{enumerate}

Thus, each vertex moves from buffer to cache, optionally oscillates between cache and stack, and finally vanishes from the cache. The interacting lifecycles of ordered buffer vertices dictate the subsets of cache vertices possible, and therefore the subgraphs of $G$ that can be parsed. Furthermore, the \textit{PushIndex} and \textit{Pop} actions construct a tree decomposition in a top-down fashion; tree bags are unique cache states $\{\eta_t\}_{t=1}^T$ and each arc is drawn by a \textit{PushIndex} action at time \textit{t} linking parent cache state $\eta_{t - 1}$ to child $\eta_{t}$.

\subsection{Simplified Transitions for NLG}
\label{sec:simp}
Since the goal for the generation task is to transform graphs into their sentences, we can prune the set of parsing actions without loss of utility:
\begin{itemize}
  \item Remove \textit{Arc} actions as the graph is given.
  \item Merge \textit{Shift} and \textit{PushIndex} actions as the former always precedes the latter. Refer to this joint action as \textit{Push}.
\end{itemize}
For the first change, edge building actions are redundant with the AMR graph as input. The second change simply (i) balances the number of  and pop actions and (ii) minimizes action sequence length. This way, the \textit{Push} and \textit{Pop} transitions can be seen as analogous to open and close parentheses, respectively. An AMR concept $v$ is active in `scope' from when it is pushed to the cache until it is evicted by the preceding vertex $u$ popped from the stack. Since the graph is traversed in word order, this is equivalent to visiting the tree bag for which a vertex is rightmost in the cache upon \textit{Push}, touring all of its children, and returning to this root prior to \textit{Pop}.

\section{Action-Conditioned Decoder}
\label{sec:ac}
From the final graph encoder \cite{song2018graph} time step, the action-conditioned decoder receives per-concept hidden states $\{\mathbf{h}_i\}_{i = 1}^n$, where $n = |G|$. As in previous systems, the decoder emits target-side tokens using an LSTM for temporal context. However, the present decoder uses hard instead of soft attention on a sequence of parsed AMR concept states.

The model attends to concepts in the order in which a predicted parser action sequence processes them. This dependency of the English sequence $\mathbf{w}$ on the actions $\mathbf{a}$ can be expressed as $P(\mathbf{a}, \mathbf{w}) = P(\mathbf{a})P(\mathbf{w} | \mathbf{a})$, where
\begin{align*}
  P(\mathbf{a}) &= \prod_{i = 1}^{2n} P(a_i | \mathbf{a}_{<i}),\\
  P(\mathbf{w} | \mathbf{a}) &= \prod_{j = 1}^{m} P(w_j | \mathbf{w}_{<j}, \mathbf{a}).
\end{align*}

Each of the above probabilities is modeled using separate action and English LSTMs, respectively. For both the action and English phases, the recurrent state update is computed as
\begin{align*}
  \mathbf{l}_t, \mathbf{s}_t &= \LSTM{\mathbf{s}_{t - 1}, \mathbf{x}_t},\\
  \mathbf{x}_t &= W_x[\mathbf{e}_{t - 1}; \mathbf{c}_t] + \mathbf{b}_x,
\end{align*}
where $\mathbf{l}_t, \mathbf{s}_t$ is the LSTM cell and hidden state pair, $\mathbf{e}_{t - 1}$ is the embedding of the previous token and $\mathbf{c}_t$ is a context vector.

The final output distribution is computed as
\begin{align*}
  \mathbf{f}_t &= \softmax{W_f[\mathbf{l}_t; \mathbf{c}_t] + \mathbf{b}_f},
\end{align*}
which applies to both $\mathbf{a}$ and $\mathbf{w}$ sequences. For clarity, action decoder time steps will be indexed by $i$ and English by $j$.

\subsection{Action Decoding}
\label{sec:ad-c}
During action decoding $\mathbf{c}_{i}$ receives features from the current cache, concatenating the embeddings of the rightmost cache element $\eta[k]$ and topmost stack element $\sigma[-1]$ with the previous decoder hidden state:
\begin{align*}
  \mathbf{c}_{i} &= W_a[\mathbf{h}_{\sigma_{i - 1}[-1]}; \mathbf{h}_{\eta_{i - 1}[k]}; \mathbf{s}_{i - 1}] + \mathbf{b}_a.
\end{align*}

The top row of Figure \ref{fig:ad} shows components of $\mathbf{c}_i$ involved in update of action LSTM state $s_{i}$.

\subsubsection{Buffer and Cache Index Prediction} \label{sec:bci-c} In NLG, the input to the parser is an unordered set rather than a sequence of buffer vertices. For each \textit{Push} action in $\mathbf{a}$, the parser must select which element in $\beta_{i - 1}$ enters the cache and which index of $\eta_{i - 1}$ to evict.

\begin{figure}[ht]
  \centering
  \begin{tikzpicture}[scale=0.95,
    main2/.style={red,align=left},
    main3/.style={draw,inner sep=2.5pt,rounded corners,align=left}]
    \tikzstyle{every node}=[font=\small]
    \draw (-0.5,3.25) node {\textit{Push}};
    \draw (0.9,3.25) node {\textit{Pop}};
    \draw (2.25,3.25) node {\textit{Push}};

    \draw (-0.5,2.75) node (1d) {$a_{i - 1}$};
    \draw (0.9,2.75) node (2d) {$a_i$};
    \draw (2.25,2.75) node (3d) {$a_{i + 1}$};

    \draw (-1.75,2) node (1c) {$\ldots$};
    \draw (-0.5,2) node[main] (5c) {$s_{i - 1}$};
    \draw (0.9,2) node[main] (2c) {$s_{i}$};
    \draw (2.25,2) node[main] (3c) {$s_{i + 1}$};
    \draw (3.45,2) node (4c) {$\ldots$};

    \draw (-1.75,.25) node[main3] (5b) {$\eta_{i - 2}$\\$\sigma_{i - 2}$};
    \draw (-0.5,.25) node[main3] (1b) {$\eta_{i - 1}$\\$\sigma_{i - 1}$};
    \draw (0.9,.25) node[main3] (2b) {$\eta_i$\\$\sigma_i$};
    \draw (2.25,.25) node[main3] (0b) {$\eta_{i + 1}$\\$\sigma_{i + 1}$};

    \draw (-0.5,-.75) node[main2] (2f) {$\pi_{{\beta}_{i - 1}}$\\$\pi_{{\eta}_{i - 1}}$};
    \draw (2.25,-.75) node[main2] (4f) {$\pi_{{\beta}_{i + 1}}$\\$\pi_{{\eta}_{i + 1}}$};

    \path[->]
      (5c) edge (1d)
      (2c) edge (2d)
      (3c) edge (3d)
      (1c) edge (5c)
      (5c) edge (2c)
      (2c) edge (3c)
      (3c) edge (4c);
    \path[red,->]
      (5c) edge node[right] {$\mathbf{i}_{\beta,\eta}$} (1b)
      (3c) edge node[right] {$\mathbf{i}_{\beta,\eta}$} (0b)
      (5b) edge (1b)
      (2b) edge (0b);
    \path[dashed, ->]
      (5b) edge (5c)
      (1b) edge (2c)
      (2b) edge (3c);
  \end{tikzpicture}
  \caption{Action decoding using parser context. Dashed arrows indicate extraction of $\mathbf{h}_\eta$ and $\mathbf{h}_\sigma$. Red arrows signify updates to parser configuration, with buffer and cache indices predicted per \textit{Push} action.}
  \label{fig:ad}
\end{figure}
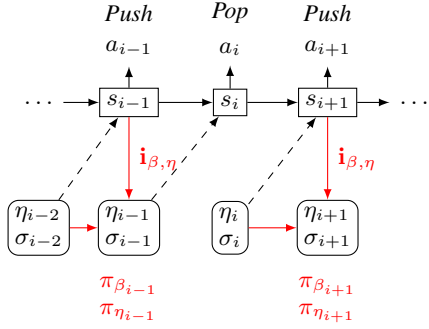

The model is trained to predict these with respect to oracle-derived index sequences, denoted $\pi^*_\beta$ and $\pi^*_\eta$. It does so by learning bilinear mappings between each data structure and the previous LSTM decoder state $\mathbf{s}_{i - 1}$ (Figure \ref{fig:ad}, bottom row):
\begin{align*}
  \mathbf{i}_{\beta_i} &= \softmax{B_{i - 1}U_\beta\mathbf{s}_i},\\
  \mathbf{i}_{\eta_i} &= \softmax{C_{i - 1}U_\eta\mathbf{s}_i},
\end{align*}
where $B_{i - 1} \in \mathbb{R}^{|\beta| \times d}$ and $C_{i - 1} \in \mathbb{R}^{|\eta| \times d}$ contain $d$-dimensional embeddings at time $i$ for elements in the buffer and cache, respectively.

\subsection{English Decoding}
The English decoder is trained to output $\mathbf{w}$ conditioned on $\pi_\beta$, or the predicted order in which buffer vertices were shifted to the cache in $\mathbf{a}$. During training, the model attends to vertices in the gold $\pi^*_\beta$ based on JAMR alignments between English spans and AMR concepts. Since the English decoder does not know length $m$ of $\mathbf{w}$ initially, it must also model concept-wise attention over sequence $\pi_\beta$.

\subsubsection{Concept-Word Alignment}
\label{sec:cw}
To align $\pi_\beta$ with $\mathbf{w}$, the model predicts progress of a pointer $p \in [1, n]$ along the former sequence. Concretely, it predicts binary \textit{increment} sequence $\mathbf{r}$ of length $m$ indicating whether $p_j$ at time $j \in [1, m]$ was incremented prior to generation of English token $w_j$. Sequence $\mathbf{r}$ is generated using bilinear mapping
\begin{align*}
  \mathbf{r}_j &= \softmax{R_jU_r\mathbf{s}_{j - 1}},
\end{align*}
where $R_j \in \mathbb{R}^{2 \times d}$ contain embeddings for current $\pi_\beta[p_j]$ and next $\pi_\beta[p_j + 1]$ concepts for corresponding positions $0$ and $1$.

\subsubsection{Cache Hard Attention}
The decoder uses increment sequence $\mathbf{r}$ to compute pointer $p_j = \sum_{k = 1}^j \argmax_l\mathbf{r}_{k, l}$ for indicator $l \in \{0, 1\}$. In this way, it attends to the rightmost cache element $\pi_\beta[p_j] = \eta_{t_e}[k]$ for parsing time step $t_e$ at which it entered the cache. For English decoding, input to the context vector $\mathbf{c}_{j}$ is limited to $\mathbf{h}_{\eta_{t_e}[k]}$ to focus attention on this newest cache addition:
\begin{align*}
  \mathbf{c}_{j} &= W_e\mathbf{h}_{\eta_{t_e}[k]} + \mathbf{b}_e,
\end{align*}
where $t_e$ refers to the parser time step determined by concept $\pi_\beta[p_j]$.

\subsection{Learning}
\label{sec:cond-l}
The total loss for the action and English decoders is combined cross-entropy loss over all predicted sequences:
\begin{multline*}
  \mathcal{L}_c = \sum_{i = 1,\:i^{\prime}: a_i^* = \textit{Push}}^{2n}\big[\mathcal{L}(a_{i}^*, a_{i}) + \mathcal{L}(i_{\beta_{i^\prime}}^*, i_{\beta_{i^\prime}})\:+\\
                 \mathcal{L}(i_{\eta_{i^\prime}}^*, i_{\eta_{i^\prime}})\big] + \sum_{j = 1}^m\big[\mathcal{L}(w_j^*, w_j) + \mathcal{L}(r_{j}^*, r_{j})\big],
\end{multline*}
where $i^\prime$ selects parsing steps for which the buffer and cache indices are predicted.

\subsection{Inference}
As the English decoder is conditioned action sequence generation, we use separate beam search routines for $\mathbf{a}$ and $\mathbf{w}$. This ensures that the buffer index sequence $\pi_\beta$ is defined prior to English generation for prediction of increment sequence $\mathbf{r}$. An action hypothesis is complete when the parser can neither push nor pop (i.e., empty buffer and stack).

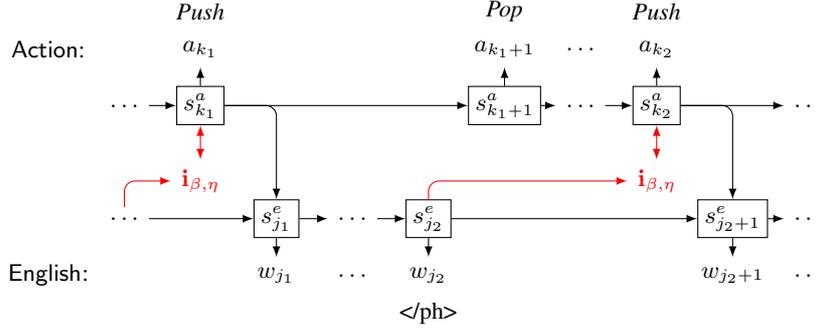
\begin{figure*}[ht]
  \centering
  \begin{tikzpicture}
    \tikzstyle{every node}=[font=\small]
    \draw (-3,.25) node {$\mathsf{Action}$:};
    \draw (-2,-.5) node (s1) {$\ldots$};
    \draw (-2,-2) node (s2) {$\ldots$};
    \draw (-1,0.75) node {\textit{Push}};
    \draw (-1,.25) node (a3) {$a_{k_1}$};
    \draw (-1,-.5) node[main] (s3) {$s_{k_1}^a$};
    \draw (-1,-1.5) node[red] (i1) {$\mathbf{i}_{\beta,\eta}$};

    \draw (0,-2) node[main] (s4) {$s_{j_1}^e$};
    \draw (1,-2) node (s5) {$\ldots$};
    \draw (2,-2) node[main] (s6) {$s_{j_2}^e$};

    \draw (0,-2.75) node (w1) {$w_{j_1}$};
    \draw (1,-2.75) node {$\ldots$};
    \draw (2,-2.75) node (w3) {$w_{j_2}$};
    \draw (-3,-2.75) node {$\mathsf{English}$:};
    \draw (2,-3.25) node {</ph>};

    \draw (3,.75) node {\textit{Pop}};
    \draw (3,.25) node (a4) {$a_{k_1 + 1}$};
    \draw (4,.25) node {$\ldots$};
    \draw (5,0.75) node {\textit{Push}};
    \draw (5,-1.5) node[red] (i2) {$\mathbf{i}_{\beta,\eta}$};
    \draw (5,.25) node (a6) {$a_{k_2}$};
    \draw (3,-.5) node[main] (s7) {$s_{k_1 + 1}^a$};
    \draw (4,-.5) node (s8) {$\ldots$};
    \draw (5,-.5) node[main] (s9) {$s_{k_2}^a$};

    \draw (7,-.5) node (s12) {$\ldots$};
    \draw (6,-2) node[main] (s10) {$s_{j_2 + 1}^e$};
    \draw (7,-2) node (s11) {$\ldots$};
    \draw (6,-2.75) node (w4) {$w_{j_2 + 1}$};
    \draw (7,-2.75) node (w5) {$\ldots$};

    \tikzset{-|/.style={to path={-| (\tikztotarget)}, rounded corners=4pt},
    |-/.style={to path={|- (\tikztotarget)}, rounded corners=4pt}}
    \path[->]
      (s1) edge (s3)
      (s2) edge (s4)
      (s3) edge (a3)
      (s4) edge (w1)
      (s6) edge (w3)
      (s3) edge (s7)
      (s4) edge (s5)
      (s5) edge (s6)
      (s7) edge (a4)
      (s9) edge (a6)
      (s7) edge (s8)
      (s8) edge (s9)
      (s10) edge (w4)
      (s10) edge (s11)
      (s6) edge (s10)
      (s9) edge (s12)
      (s3) edge[-|] (s4)
      (s9) edge[-|] (s10);
    \path[->,red]
      (s2) edge[|-] (i1)
      (s6) edge[|-] (i2);
    \path[<->,red]
      (s3) edge (i1)
      (s9) edge (i2);
  \end{tikzpicture}
  \caption{Depiction of action-word LSTM interaction in the joint decoder. For clarity, step-wise parser configuration is absorbed into action LSTM decoder states. Red arrows capture prediction of $\mathbf{i}_\beta$ and $\mathbf{i}_\eta$, which are then used to update the parser state upon \textit{Push}. This updated state is used for cache hard attention by the English LSTM.}
  \label{fig:jd}
\end{figure*}

\section{Joint Action-Word Decoder}
\label{sec:jaw}
The decoder produces a sequence of interleaved words $w_{1:m}$ and actions $a_{1:2n}$ according to JAMR alignment. Each English span is preceded by a \textit{Push} action for its aligned concept and terminates in an end of phrase symbol, </ph>.  Below is a target string for the parser run in Table~\ref{tab:sample-run}, where the action subscripts refer to the traversal order $\pi_\beta = [\texttt{center}, \texttt{formal}, \texttt{open-01}, \texttt{date-entity},\\\texttt{2009}]$.
\begin{description}
  \item \textit{Push\textsubscript{1}} the center will </ph> \textit{Push\textsubscript{2}} formally </ph> \textit{Push\textsubscript{3}} open in </ph> \textit{Push\textsubscript{4}} \textit{Push\textsubscript{5}} 2009 . </ph> \textit{Pop\textsubscript{5}} \textit{Pop\textsubscript{4}} \textit{Pop\textsubscript{3}} \textit{Pop\textsubscript{2}} \textit{Pop\textsubscript{1}}
\end{description}

Let $\mathbf{y} = y_{1:(2n + m)}$ denote the merged actions $\mathbf{a}$ and English words $\mathbf{w}$. The model iteratively generates actions given previous English spans and vice versa. It maximizes the joint probability
\begin{align*}
  P(\bar{\mathbf{a}}, \bar{\mathbf{w}}) &= \prod_{i = 1}^{2n + m} P(y_i | \bar{\mathbf{a}}_{<i}, \bar{\mathbf{w}}_{<i}),\\
  y_i &= \begin{cases}
    \bar{a}_i,& \text{if } y_{i - 1} \in \{\text{\it{Pop}}, \text{</ph>}\}\\
    \bar{w}_i,& \text{otherwise,}
  \end{cases}
\end{align*}
where $\bar{\mathbf{a}}$ and $\bar{\mathbf{w}}$ are length-normalized versions of $\mathbf{a}$ and $\mathbf{w}$ to accept indices $i \in [1, 2n + m]$. Clearly, $\bar{a}_i$ and $\bar{w}_i$ are defined only when the model is in the appropriate action or English phase at time $i$.

The joint decoder produces output similarly to the conditional decoder (Section \ref{sec:ac}), as it uses separate action and English LSTMs corresponding to its generation phase. However, since jointly decoded English spans are implicitly aligned to the AMR concept for their preceding \textit{Push} action, the pointer $p$ and increment sequence $\mathbf{r}$ (Section \ref{sec:cw}) are removed. In addition, the action LSTM can now receive context from previously generated English spans.

\subsection{Action Decoding}
Context vector $\mathbf{c}_i$ for the action LSTM is composed of concatenated stack, cache, and decoder state features. In comparison to Section \ref{sec:ad-c}, the stack and cache embeddings are averaged over elements in each data structure:
\begin{align*}
  \mathbf{c}_{i} &= W_a[\mathbf{h}_{\avg{\sigma}}; \mathbf{h}_{\avg{\eta}}; \mathbf{s}_{i - 1}] + \mathbf{b}_a.
\end{align*}
This change is intended to provide a more global view of parser context, as opposed to only the stack top and rightmost cache elements.

\subsubsection{Buffer and Cache Index Prediction}
The index sequences $\mathbf{i}_{\beta_i}$ and $\mathbf{i}_{\eta_i}$ are predicted using the same form as in Section \ref{sec:bci-c}. However, the respective embedding matrices $B_{i - 1}$ and $C_{i - 1}$ are now augmented with buffer-cache subgraph edge embeddings. In addition, decoder state $\mathbf{s}_{i - 1}$ is concatenated with action LSTM state $\mathbf{s}_{i^\prime}^a$, where $i^\prime = \max{\{i: \bar{a}_i = \textit{Push}\}}$. The updated index predictions can be written as
\begin{align*}
  \mathbf{i}_{\beta_i} &= \softmax{B_{i - 1}^\prime U_\beta[\mathbf{s}_i;\mathbf{s}_{i^\prime}^a]},\\
\mathbf{i}_{\eta_i} &= \softmax{C_{i - 1}^\prime U_\eta[\mathbf{s}_i;\mathbf{s}_{i^\prime}^a]}.
\end{align*}

The modified buffer embedding matrix $B_{i - 1}^\prime \in \mathbb{R}^{|\beta| \times (d + 2d^\prime)}$, where $d$ is vertex and $d^\prime$ is edge label embedding dimensionality. The embedding for non-empty index $j \in [1, |\beta|]$ is
\begin{align*}
  B_{i - 1,j}^\prime &= [\mathbf{h}_{\beta[j]}; \mathbf{e}_\text{i}; \mathbf{e}_\text{o}],
\end{align*}
where AMR edge label embeddings
\begin{align*}
  \mathbf{e}_\text{i} = \sum_{j \in \mathcal{N}(k)}\mathbf{e}_{k, j}^\intercal / |\eta|,\;
  \mathbf{e}_\text{o} = \sum_{k \in \mathcal{N}(j)} \mathbf{e}_{j, k} / |\eta|.
\end{align*}
Note that $k \in \eta$ and $\mathcal{N}(\cdot)$ contains outgoing vertices in the bipartite subgraph between buffer and cache (or vice versa). The updated cache embedding matrix $C_{i - 1}^\prime$ is defined analogously.

Subgraph edge embeddings are used to provide the decoder with localized graph structure between buffer and cache. Buffer elements most strongly connected to the cache would likely be highly scored. In contrast, cache elements with fewer buffer edges may be chosen for eviction.

\subsection{English Decoding}
To generate each valid English word $\bar{\mathbf{w}}_i$ the decoder uses hard attention on the most recently pushed element $\eta_i[k]$. The English span generation phase is triggered upon $\bar{\mathbf{a}}_{i^\prime} = \textit{Push}$ for $i^\prime < i$, so this element corresponds to $\mathbf{i}_{\beta_{i^\prime}}^*$ (or $\argmax{\mathbf{i}_{\beta_{i^\prime}}}$ at test time). Thus, the English LSTM context vector is simply $\mathbf{c}_i = W_e\mathbf{h}_{\eta_{i^\prime}[k]} + \mathbf{b}_e$.

To incorporate context from the action LSTM, the English output distribution is modified to contain the preceding action hidden state $\mathbf{s}_{i^\prime}^a$:
\begin{align*}
  \mathbf{f}_i &= \softmax{W_f[\mathbf{l}_i; \mathbf{c}_i; \mathbf{s}_{i^\prime}^a] + \mathbf{b}_f}.
\end{align*}
This added state encapsulates the sequence of buffer elements pushed to the cache and may aid in sequential English word generation as well.

\subsection{Learning}
Let virtual sequence $\mathbf{y} = \bar{\mathbf{a}} \cup \bar{\mathbf{w}}$, where $|\mathbf{y}| = 2n + m$, represent interleaved actions $\mathbf{a}$ and English tokens $\mathbf{w}$. The full loss format follows that of Section \ref{sec:cond-l}:
\begin{multline*}
  \mathcal{L}_j = \sum_{i = 1,\:i^{\prime}: a_i^* = \textit{Push}}^{2n + m} \big[\mathcal{L}(y_i^*, y_i) + \mathcal{L}(i_{\beta_{i^\prime}}^*, i_{\beta_{i^\prime}})\:+\\
                 \mathcal{L}(i_{\eta_{i^\prime}}^*, i_{\eta_{i^\prime}})\big].
\end{multline*}
Since decoding during inference completes only when the parser configuration reaches a terminal state, $\mathbf{y}$ contains all $2n$ actions during training.

\subsection{Inference}
With the alternation between actions and English spans, beam search likewise must switch between its two generation phases. Each sequence always starts in the action phase, and phase switching is deterministic upon $\bar{a}_i = \textit{Push}$ or $\bar{w}_i = \text{</ph>}$.

Thus, beam search occurs primarily over actions, with hypotheses ending in $\bar{a}_i = \textit{Push}$ entering into a subroutine of English beam search. Upon completion of the latter, an action hypothesis is created for each new English one. The original `parent' action hypothesis is destroyed and all remaining action hypotheses are rescored and sorted with respect to both $\mathbf{a}$ and $\mathbf{w}$. More precisely, the score of each action hypothesis sums over log probabilities of both its action and English sequences.

\section{Experiments}
We experiment on the most recently released AMR corpus, LDC2017T10, containing $36,521$ training pairs and the same development ($1,368$) and test ($1,371$) sets as LDC2015E86. To avoid compounded noise from extracted parser actions, no automatically annotated external data are used for training.

\subsection{Preprocessing}
Following \citet{song2018graph}, the only modifications made to the input AMRs affect the concept labels: conversion to lowercase and removal of PropBank sense IDs (e.g., \texttt{run-02} $\to$ \texttt{run}). The latter improves label overlap with the pretrained word embeddings shared with English tokens.

We use the oracle extraction algorithm of \citet{peng2018amr} to parse the gold sentence-AMR pairs. However, we do not categorize the AMR graphs as fine granularity concept attributes are desirable in NLG. We then apply the reduced transition set described in Section \ref{sec:simp} to these action sequences, while recording the evicted cache index for each \textit{Push} action.

As previously mentioned we use JAMR on the gold data to align word spans and AMR concepts. To create interleaved sequences $\mathbf{y}$, unaligned English tokens are included in the preceding AMR concept's span according to $\pi_\beta^*$.

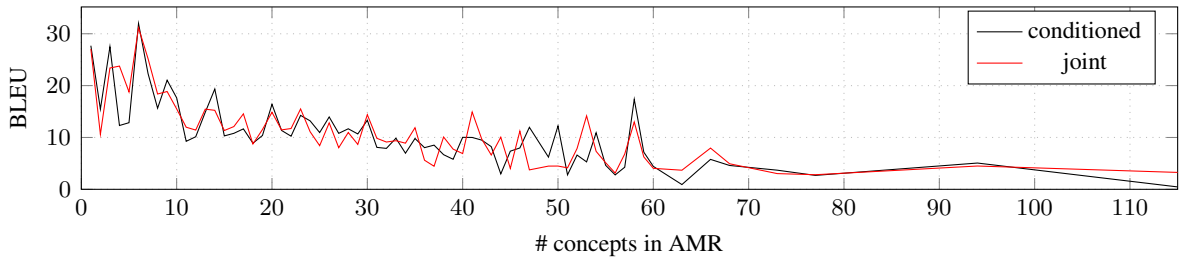
\begin{figure*}[ht]
  \centering
  \begin{tikzpicture}
    \tikzstyle{every node}=[font=\small]
    \begin{axis}[
      grid=major,
      grid style={dotted},
      xmin=0,
      xmax=115,
      ymin=0,
      xlabel={\# concepts in AMR},
      ylabel=BLEU,
      height=0.25\textwidth,
      width=\textwidth]
      \addplot[no marks] table [x=n,y=b,col sep=comma] {bin_bleu_c.csv};
      \addplot+[no marks] table [x=n,y=b,col sep=comma] {bin_bleu_j.csv};
      \legend{conditioned,joint}
    \end{axis}
  \end{tikzpicture}
  \caption{Degradation of test BLEU score over AMR graph size.}
  \label{fig:deg}
\end{figure*}

\subsection{Setup}
The word embeddings are initialized with $300$-dimensional GloVe embeddings \cite{pennington2014glove} from the Common Crawl and are fixed during training. This embedding vocabulary consists of both English tokens and AMR concept labels. Before being passed on to the decoder, encoder hidden states are concatenated with word embeddings for concept labels.

All model variants are implemented in TensorFlow. Instances of action and English LSTMs are single layer with hidden state size $512$, which is also the size of intermediate hidden vectors in the model. For SGD, \textit{Adam} \cite{kingma2014adam} is used with a learning rate of \num{1e-3}. Other encoder-related parameters are set according to the gold data settings from \citet{song2018graph}.

Hyperparameters are tuned on the development set of LDC2017T10. To correct for overly short English sequences, we use a constant length reward $\epsilon$ multiplied by predicted English sequence length when evaluating beam search hypotheses. We report results using the BLEU \cite{papineni2002bleu} evaluation metric.

\subsection{Results}
We first report results on the full test set, then investigate the negative effects of graph size on sentence generation quality.
\begin{table}[h]
  \centering
  \begin{tabular}{@{}l r r r@{}}
    \toprule
    & & \textbf{dev} & \textbf{test}\\
    \midrule
    \multicolumn{2}{l}{conditioned} & $12.50$ & $11.71$\\
    & $+$ oracle $\pi_\beta^*$ & $19.71$ & $19.51$\\
    \multicolumn{2}{l}{joint} & $12.86$ & $11.97$\\
    & $+$ oracle $\pi_\beta^*$ & $16.78$ & $16.32$\\
    \bottomrule
  \end{tabular}
  \caption{Overall BLEU, along with scores using oracle buffer index sequences.}
  \label{tab:bleu}
\end{table}

Table \ref{tab:bleu} compares BLEU across model variants for both development and test splits. The conditioned model clearly outperforms the joint model, despite the ability for the latter's action decoder to receive context from its English LSTM. Perhaps the conditioned model is better able to learn parser behavior as its action LSTM focuses exclusively on parser context.

Regardless, the uncompetitive BLEU scores suggest several possible failure points. The main limitation appears to be poor prediction of $\pi_\beta$, or the order of AMR concepts pushed to the buffer. Table \ref{tab:bleu} highlights this difference between using predicted $\pi_\beta$ and oracle $\pi_\beta^*$ as $\sim7$ BLEU in the conditioned case and $\sim4$ in the joint. The fact that the joint model's English decoder depends on action LSTM state in addition to parser context may explain why it is less affected by the oracle $\pi_\beta^*$.

\begin{table}[h]
  \centering
  \begin{tabular}{@{}l r r r r r@{}}
    \toprule
    & $\mathbf{w}$ & $\mathbf{a}$ & $\mathbf{i}_\beta$ & $\mathbf{i}_\eta$ & $\mathbf{r}$\\
    \midrule
    conditioned & $0.62$ & $0.95$ & $0.72$ & $0.87$ & $0.78$\\
    joint & $0.65$ & $0.93$ & $0.75$ & $0.88$ & --\\
    \bottomrule
  \end{tabular}
  \caption{Best per-sequence development set accuracies during training.}
  \label{tab:accu}
\end{table}

Another issue may be error propagation in parser-related sequences. Figure \ref{fig:deg} shows the negative effect of AMR size on generation quality. The models only exceed $20$ BLEU on AMR with fewer than $10$ concepts. Presumably, parser decisions become more difficult with longer action and buffer index sequences. The number of possible action sequences $\mathbf{a}$ is given by Catalan number $C_n = \binom{2n}{n}/(n + 1)$. The $n!$ possible buffer index sequences $\pi_\beta$ presents a similarly complex output space. Coupled together, errors in either sequence can lead to rapid divergence in train and test settings. A model trained on gold parser states will likely suffer when forced to condition its English generation on unseen parser states.

From Table \ref{tab:accu} it is apparent that both model variants achieve relatively high accuracies during training. This is likely due to reliance on gold sequence histories from teacher forcing.

\section{Conclusion}
We introduce two variants of AMR-to-text generation models that produce semantic parsing actions either before or during English production. Both model variants use a recurrent graph encoder to learn concept-level states from AMR structure. Utilizing parser state was expected to (i) model word-order traversal of the input AMR and (ii) explicitly align English spans to AMR concepts using hard attention. Due to train-test divergence on parser sequence prediction, the models were unable to attain competitive performance.

\bibliographystyle{acl_natbib}
\bibliography{emnlp-ijcnlp-2019}
\end{document}